%% file: main.tex
\pdfoutput=1

\PassOptionsToPackage{sort,sort&compress}{natbib}

\documentclass[11pt]{article}
\usepackage[final]{acl}
\usepackage[utf8]{inputenc}
\usepackage{graphicx}
\usepackage{svg}
\usepackage{times}
\usepackage{latexsym}
\usepackage[T1]{fontenc}
\usepackage{amsmath,amsfonts,amssymb,bm}
\usepackage{microtype}
\usepackage{inconsolata}

\usepackage[capitalise,nameinlink,noabbrev]{cleveref}
\usepackage{tabularx}
\usepackage{booktabs}
\usepackage{siunitx}
\usepackage{caption}
\usepackage{multirow}
\usepackage{wrapfig}
\usepackage{pifont,fontawesome}
\usepackage{import}
\usepackage{longtable}
\usepackage{etoolbox}
\usepackage{placeins}
\usepackage{arydshln}    

\usepackage{subcaption}

\newcommand{\code}[1]{\texttt{#1}}

\newcommand{\crefsubtable}[2]{\hyperref[#1]{\cref*{#1}#2}}

\usepackage{ulem}
\usepackage{pgffor}


\robustify\bfseries
\robustify\uline

\makeatletter
\patchcmd{\@@setcref}         {??}{\color{orange} ??}{}{}
\patchcmd{\@@setcref}         {??}{\color{orange} ??}{}{}
\patchcmd{\@@setcrefrange}    {??}{\color{orange} ??}{}{}
\patchcmd{\@@setcrefrange}    {??}{\color{orange} ??}{}{}
\patchcmd{\@@setcrefrange}    {??}{\color{orange} ??}{}{}
\patchcmd{\@@setcrefrange}    {??}{\color{orange} ??}{}{}
\patchcmd{\@@setcrefrange}    {??}{\color{orange} ??}{}{}
\patchcmd{\@@setcrefrange}    {??}{\color{orange} ??}{}{}
\patchcmd{\@@setnamecref}     {??}{\color{orange} ??}{}{}
\patchcmd{\@@setnamecref}     {??}{\color{orange} ??}{}{}
\patchcmd{\@@setcpageref}     {??}{\color{orange} ??}{}{}
\patchcmd{\@@setcpageref}     {??}{\color{orange} ??}{}{}
\patchcmd{\@@setcpagerefrange}{??}{\color{orange} ??}{}{}
\patchcmd{\@@setcpagerefrange}{??}{\color{orange} ??}{}{}
\patchcmd{\@@setcpagerefrange}{??}{\color{orange} ??}{}{}
\patchcmd{\@@setcpagerefrange}{??}{\color{orange} ??}{}{}
\patchcmd{\@@setcpagerefrange}{??}{\color{orange} ??}{}{}
\patchcmd{\@@cref}            {??}{\color{orange} ??}{}{}
\makeatother

\crefformat{section}{#2§#1#3}
\crefformat{subsection}{#2§#1#3}

\usepackage{enumitem}
\setlist{itemsep=0pt,parsep=2pt, leftmargin=\labelwidth}

\input{content/math_commands.tex} 

\title{FOSSIL: Harnessing Feedback on Suboptimal Samples for Data-Efficient Generalisation with Imitation Learning for Embodied Vision-and-Language Tasks}

\author{Sabrina McCallum\textsuperscript{1,2}, Amit Parekh\textsuperscript{2}, Alessandro Suglia\textsuperscript{1} \\
\textsuperscript{1}University of Edinburgh \\
\textsuperscript{2}Heriot-Watt University \\
\texttt{\{s.mccallum-exner, asuglia\}@ed.ac.uk, a.parekh@hw.ac.uk}
}

\usepackage{xspace,abbrevs}

\newabbrev{\MSA}{\textit{My Super Awesome Abbreviation}}

\makeatletter
\renewcommand\maybe@space@{%
  \maybe@ictrue 
  \expandafter   \@tfor
    \expandafter \reserved@a
    \expandafter :%
    \expandafter =%
                 \nospacelist
                 \do \t@st@ic
  \ifmaybe@ic 
    \space
  \fi
}
\makeatother

\begin{document}
\maketitle

\begin{abstract}
Current approaches to embodied AI tend to learn policies from expert demonstrations. However, without a mechanism to evaluate the quality of demonstrated actions, they are limited to learning from optimal behaviour, or they risk replicating errors and inefficiencies. While reinforcement learning offers one alternative, the associated exploration typically results in sacrificing data efficiency. This work explores how agents trained with imitation learning can learn robust representations from both optimal and suboptimal demonstrations when given access to constructive language feedback as a means to contextualise different modes of behaviour. We directly provide language feedback embeddings as part of the input sequence into a Transformer-based policy, and optionally complement the traditional next action prediction objective with auxiliary self-supervised learning objectives for feedback prediction. We test our approach on a range of embodied Vision-and-Language tasks in our custom \textsc{BabyAI-XGen} environment and show significant improvements in agents' compositional generalisation abilities and robustness, suggesting that our data-efficient method allows models to successfully convert suboptimal behaviour into learning opportunities. Overall, our results suggest that language feedback is a competitive and intuitive alternative to intermediate scalar rewards for language-specified embodied tasks.
\end{abstract}


\section{Introduction}

\subimport{figures}{overview}

Embodied AI presents a significant advancement in artificial intelligence, emphasising the importance of physical embodiment and the resulting ability to interact directly with the environment to accomplish goals. By perceiving and reasoning about the effect of their actions on their surroundings, embodied AI systems gain access to rich learning signals, which enable more robust and context-aware models of the world \citep{deitke2022retrospectives}. For such systems to be useful across real-world deployment scenarios, they must be able to translate language instructions into meaningful sequences of actions. Embodied agents that understand language not only unlock an additional learning modality, but also enable richer, situational representations of behaviour by grounding language into additional sensory modalities such as vision. This facilitates the creation of language-guided pixel-based policies for Embodied AI (e.g., \citealp{Team2024OctoAO}).

While imitation learning (IL) methods are commonly used to train such agents, it is typically assumed that the underlying demonstration data reflects optimal or near-optimal behaviour \citep{Min2022DontCT}. This presents three key limitations. First, models trained exclusively on optimal trajectories may learn that there is exactly one valid solution for a given task.

Second, such models never encounter recoverable errors and their corrections. Third, IL lacks mechanisms to evaluate the quality of actions when faced with multiple valid behavioural modes for the same observation. In contrast, reinforcement learning (RL) algorithms explicitly incorporate feedback in the form of scalar rewards to distinguish between varied, more or less optimal behaviours \citep{sutton2018reinforcement, levine2020offline}. However, the associated exploration in RL also means that it is typically less sample-efficient than IL. This is exacerbated for sparse rewards, which is typically the case in embodied AI. 

In this work, we investigate to what extent Transformer-based IL policies can benefit from the inclusion of suboptimal demonstrations in the training data when the effect of actions is contextualised with constructive language feedback, turning mistakes into learning opportunities. We focus on this family of models due to their ability to better represent multimodal inputs and long input sequences that are common in Embodied AI tasks, and to align our work with the foundational architecture and training regimen used for current Vision-Language-Action (VLA) models in robotics \citep{Shukor2025SmolVLAAV, Nvidia2025GR00TNA, Black20240AV, Kim2024OpenVLAAO}. We test whether our approach can yield improved generalisation capabilities of language-guided IL policies, while simultaneously ensuring sample-efficient learning. 

We introduce a training regime where we systematically augment the training data with suboptimal variations of the base demonstrations. The use of language feedback is intuitive in the light of possible real-world deployment scenarios for embodied AI, and mirrors the modality of language instructions. In addition, we test if language feedback is a competitive substitute or suitable complement for shaped scalar rewards when conditioning IL on feedback. We further test the efficacy of a self-supervised auxiliary task in the form of feedback prediction. The motivation for introducing the feedback prediction objectives is two-fold: 1. To encourage more robust representations of actions, and 2. To give the model the ability to tap into an internal world model of the consequences of actions when no feedback is provided at inference. To allow us to test fine-grained compositional generalisation capabilities systematically, we develop BabyAI-XGen, a modified version of the BabyAI \citep{ChevalierBoisvert2018BabyAIAP} environment, bridging the gap between procedural generation and granular control over task configurations and environment parameters. 

Our findings indicate that Transformer-based IL policies that are trained from scratch with language feedback-conditioned suboptimal demonstrations generalise significantly better in compositional tasks than a baseline trained only on optimal trajectories, by turning mistakes and inefficient strategies into learning opportunities. We find this effect is consistent across different amounts of training data.

Notably, we observe that language feedback and scalar rewards provided with similar frequency yield comparable performance, offering practical flexibility depending on which signal is easier to provide. While combining language feedback and scalar rewards does not significantly improve task success rates, it does improve robustness to input perturbations during inference, indicating their complementary strengths. The same is observed for the auxiliary prediction task. Lastly, we present further evidence for the sample efficiency of our method by demonstrating that an online RL baseline trained with PPO \citep{Schulman2017ProximalPO} and using the same reward function achieves near-random generalisation performance when trained on equivalent numbers of data points. 

In summary, the contributions of this work are as follows:

\begin{itemize}[topsep=2pt]
    \item We present \texttt{FOSSIL} (\textbf{F}eedback on \textbf{S}uboptimal \textbf{S}amples in \textbf{I}mitation \textbf{L}earning), a framework that leverages feedback to unlock the learning potential in suboptimal demonstrations for IL.
    \item We release \texttt{BabyAI-XGen}, a modified version of BabyAI for the procedural generation of custom tasks for compositional generalisation.
    \item We define a range of evaluation settings to assess compositional generalisation, robustness to perturbations and data efficiency, and present results that suggest that language feedback and shaped scalar rewards can be equally effective.
\end{itemize}


\section{Related work}

\paragraph{Language-guided embodied AI.} We consider our work in the context of language-guided embodied AI, specifically, mobile manipulation tasks. Compared with manipulation-only tasks \citep[etc.]{mees2022calvin, jiang2023vima}, and navigation-only tasks \citep{anderson2018vision}, the combination of navigation and object interaction creates additional degrees of freedom for behaviour to deviate from the optimal solutions, and hence, more meaningful opportunities to provide feedback can arise. Existing datasets \citep{Shridhar2019ALFREDAB, Shridhar2020ALFWorldAT, Gao2022DialFREDDA, puig2018virtualhome, yenamandra2023homerobot} tend to annotate planner-generated trajectories with language instructions, which means that agents are trained to rely on optimal behaviour, while suboptimal behaviour is seen as something to mitigate, not leverage \citep{Min2022DontCT}.

\paragraph{Language as feedback.} Previous work uses LLMs as judges \citep{wu2024meta, pang2024iterative}, trains models to provide language feedback \citep{Zhong2024PolicyIU} or leverages LLMs to increase the diversity of language feedback \cite{Xi2024TeachingER} for a range of tasks. In contrast, we investigate \textit{how to utilize language feedback systematically to contextualise different modes of behaviour}, rather than on the quality or generation mode of the feedback. While definitions of feedback in previous work range widely, it broadly falls into two categories: 1) general language hints \citep{Lin2023LearningTM} or foresight \citep{Xi2024TeachingER}---which often takes the form of granular instructions---and 2) hindsight feedback \cite{Xi2024TeachingER}. We focus on a form of hindsight feedback. To reduce the number of confounding variables, we procedurally populate a multi-part template which reflects both a judgment of the action and, importantly, an explanation of the judgment. Our work is also in juxtaposition to previous work that derives a reward function from language, e.g. by computing the similarity between observations and subgoals described in language \citep{Du2023GuidingPI, Adeniji2023LanguageRM}. Instead, we use the language feedback embeddings directly as input into a Transformer-based \cite{Vaswani2017AttentionIA} policy, following work by \citet{McCallum2023IsFA} and \citet{Xi2024TeachingER}. While both are antecedents to our work, they only compare language feedback with rewards corresponding to the binary failure / success scenarios typically associated with embodied AI tasks, and disregard the interplay of the language feedback with the optimality of the trajectories in the training data, which consists either of demonstration datasets \citep{Xi2024TeachingER} or randomly generated trajectories \citep{McCallum2023IsFA}. In contrast, we facilitate a fair comparison of language feedback and rewards by aligning their frequency, and systematically construct datasets composed of both optimal and suboptimal trajectories. Another important strength of our work is that, differently from \citet{Xi2024TeachingER}, we focus on purely pixel-based policies rather than studying environments where agents have access to symbolic state representation~\footnote{In \citet{Xi2024TeachingER}'s setup, only one environment over four uses pixel-based representations.}.

\paragraph{Self-supervised auxiliary tasks for grounded representations.} While predicting the observations and rewards resulting from actions is inherent to world models \citep{Zhang2024LanguageguidedWM, Du2023GuidingPI, Hafner2023MasteringDD, Hafner2020MasteringAW, Hafner2019DreamTC}, where this ability is commonly referred to as \emph{latent imagination}, the utility of predicting tokens other than actions as self-supervised auxiliary tasks \citep{jaderberg2016reinforcement} in IL is still under-explored. Results from work on other multi-modal tasks, such as multi-modal translation \citep{Elliott2017ImaginationIM} or guessing games \citep{Suglia2020ImaginingGC}, suggest that such training objectives are a promising avenue for learning well-grounded multi-modal representations. We adapt this approach for IL for embodied instruction following tasks.

\paragraph{Compositional generalisation in embodied AI.} Previous work, such as VIMA~\cite{jiang2023vima} for tabletop manipulation, predominantly investigates \textit{Systematicity}, or the ability of models to systematically recombine known components and rules to novel combinations \citep{Hupkes2019CompositionalityDH}. However, other dimensions of compositional generalisation, such as \textit{Productivity}, remain under-explored in the context of embodied agents and interactive environments. In lieu of existing suitable settings, we design a multi-dimensional framework to evaluate the compositional generalisation abilities of models trained with our method, and facilitate this with our \textsc{BabyAI-XGen} environment.


\section{Method}\label{sec:method}

\subimport{figures}{masking-strategies-shortened}

Inspired by Decision Transformers \citep{Chen2021DecisionTR} and \texttt{Uni[MASK]} \citep{Carroll2022UniMASKUI}, we model different types of IL as sequential decision problems, with the option to condition action generation on relevant additional tokens corresponding to data elements that may accompany state-action pairs, such as instructions,  scalar rewards and/or language feedback. All our models share the same base architecture, consisting of an autoregressive Transformer backbone with a simple action prediction head, with optional additional heads to predict tokens for auxiliary tasks. \cref{fig:masking-strategies-shortened} illustrates an example configuration using our flexible architecture. The token masks for all models used in our experiments and additional possible configurations can be found in \cref{app:appendix-b}. We further deviate from the original Decision Transformer and previous work applying Decision Transformer to similar domains \citep{Xi2024TeachingER} by using Llama2 \citep{Touvron2023LLaMAOA} as our reference backbone architecture\footnote{Since we override the configuration, we do not use Llama2's pretrained weights.} to facilitate faster training and inference, as well as support learning from entire trajectories, rather than sub-trajectories of limited context lengths. Language inputs are encoded using a frozen pretrained Sentence-BERT language model \citep{Reimers2019SentenceBERTSE}, which allows us to condense sentences of arbitrary length into compact vector representations. Image observations are encoded into a single token using a simple CNN network trained concurrently with the policy. We provide full implementation details in \cref{app:appendix-b}.

\paragraph{Training Objectives and Loss Functions.} 
We follow previous work on Decision Transformers~\cite{Chen2021DecisionTR} and learn to predict actions in the environment as a next token prediction task by minimising cross-entropy loss of the actions. Beyond the standard cross-entropy loss for action prediction, we incorporate auxiliary self-supervised training objectives that involve predicting feedback signals at the next time step. Specifically, we use additional regression heads with MSE loss to predict scalar reward values and language feedback embeddings at t+1. These auxiliary losses are balanced with the primary action prediction loss through learnable weighting parameters. This multi-objective approach encourages the model to develop richer representations of action consequences, as it must anticipate not only appropriate actions but also the feedback those actions will generate, enabling better understanding of behaviour and context. Note that the predicted feedback tokens are used exclusively to compute auxiliary losses and are not fed back into the model as input. Implementation details for all losses and the loss balancing are provided in \cref{app:appendix-b}.


\section{Experimental framework}

\subsection{Evaluation settings}
We refer to the framework proposed by \citet{article} and study generalisation in embodied Vision-and-Language tasks along two main axes: (1) \textit{compositional generalisation} and (2) \textit{robustness}. For compositional generalisation, we consider scenarios of Systematicity and Productivity. For robustness, we analyse the robustness of models' goal representations, their robustness to external perturbations and adversarial or missing feedback, and the efficiency of the obtained solutions. We further test whether the ability to generalise compositionally is associated with the ability to effectively leverage the available samples, and investigate generalisation performance at different proportions of the training data, comparing our method with the IL and an online RL baseline.

\subsection{Environment: BabyAI-XGen}
We develop BabyAI-XGen\footnote{We make the full environment code and all task configurations available at \href{https://github.com/sabraaap/fossil}{github.com/sabraaap/fossil}.} to fill the gap in embodied environments capable of supporting rigorous compositional generalisation research and facilitate our evaluation settings. BabyAI-XGen is the result of various key modifications to BabyAI's \citep{ChevalierBoisvert2018BabyAIAP} backend, aimed at giving users control over a comprehensive range of environment parameters, allowing them to configure missions corresponding to distinct train and test scenarios. We outline all configuration parameters and their intended use in \cref{app:appendix-d}. 

We choose BabyAI as our starting point since, despite its apparent simplicity, it remains a popular and challenging testbed for embodied agents that understand and learn from natural language instructions \citep[etc.]{Lu2024IntelligentGS, Gu2024SEALSI, huang2024dark}. Out of a range of candidate grid world environments that support language \citep{Lin2023LearningTM, Zholus2022IGLUGS, Wang2021GroundingLT}, only BabyAI fulfils all desiderata required to facilitate our training and evaluation settings: partial observability, pixel observations, navigation with obstruction, multiple types of object interactions, the ability to chain together tasks, as well as objects which can be defined along several attribute axes. Furthermore, AgentBoard \citep{Ma2024AgentBoardAA} and AgentGym \citep{Xi2024AgentGymEL}, two recent LLM agent benchmarks, include BabyAI and rate its difficulty on par with more visually realistic 3D environments such as ALFWorld \citep{Shridhar2020ALFWorldAT}. Note that both benchmarks target text-only models and rely on the symbolic observations, while we use the pixel observations of the partial, egocentric views for a more challenging setting that requires vision and language. Importantly, BabyAI is highly reproducible, extensible and scalable, as it does not depend on an external game or physics engines, which makes the procedural generation of training data lightweight and reliable. 

Using BabyAI-XGen, we define new level configurations to directly isolate and study different types of compositional generalisation. We focus on variations of Pickup and PutNext tasks, since these require navigation and interaction. To make the setting even more challenging, we enforce a range of stricter settings, including shorter timeouts and a guarantee that missions for a given level consistently require the same skill set, such as understanding locations or temporal order in instructions, or navigating to a goal object located in a different room. This addresses a limitation associated with mission generation mechanism in the original BabyAI suite, which samples missions post-hoc based on the sampled objects, making it impossible to ensure that all missions for a given level strictly require the skill set attributed to that level. In contrast, we sample missions and construct the objects in the scene around them.

\subsection{Training dataset generation}
We generate two types of trajectory datasets for \textit{each task} used in our multi-task training setup: one with \textasciitilde12K optimal paths using BFS planning, and another composed of \textasciitilde4K optimal and \textasciitilde8K suboptimal paths with mistakes and inefficiencies\footnote{All datasets and the code for generating trajectories are publicly available at \href{https://huggingface.co/fossil-eai/datasets}{huggingface.co/fossil-eai/datasets}. and \href{https://github.com/sabraaap/fossil}{github.com/sabraaap/fossil}, respectively.}. For suboptimal trajectories, we uniformly sample from the action space and replace the planner action with probability $p$, where the subsequent planner action corrects any suboptimal behaviour. The process is depicted in \cref{fig:mixed-trajectories}.

\subimport{figures}{mixed-trajectories}

\subsection{Feedback augmentation}
We generate language feedback and scalar rewards according to the same underlying heuristics, by building on the rule-based feedback oracles proposed by \citet{McCallum2023IsFA}. This approach ensures consistent, reliable and accurate feedback generation, which is essential for our controlled study of how different feedback modalities affect learning from suboptimal trajectories. Our rule-based feedback generation pipeline leverages privileged information about environment states and task objectives to provide comprehensive coverage of relevant feedback scenarios. This systematic approach allows us to isolate the effects of feedback modality and quality without introducing variability from potential inconsistencies in model-generated feedback \citep{Xi2024TeachingER}.

\paragraph{Task feedback} We extend the oracle for ``task feedback'' from \citet{McCallum2023IsFA} to return not only \textit{positive} language feedback for \textit{desired} interactions with \textit{goal objects} that lead to the agent completing a subgoal, but additionally \textit{negative} language feedback on \textit{undesired} interactions with \textit{distractor objects}, and increase diversity and informativeness of our template-based feedback following \citet{Xi2024TeachingER}. The template, including representative examples, is illustrated in \cref{app:appendix-f}.

\paragraph{Affordance feedback} We adopt the ``rule feedback'' oracle from \citet{McCallum2023IsFA} without any changes. This feedback triggers when agents attempt actions that have no visible effect on the environment due to its affordances---such as trying to move through closed doors, open walls, or stack objects (which is not permitted in the environment).

\paragraph{Shaped Rewards.} We apply reward shaping to BabyAI's original binary reward function and derive intermediate scalar rewards by leveraging the success and action failure detection mechanisms in the feedback oracles. Note that while our rewards may appear dense relative to the original binary rewards, they are comparably sparse in the context of typical RL problems \citep[etc.]{bellemare13arcade, Tassa2018DeepMindCS, wydmuch2018vizdoom, Cobbe2019LeveragingPG}. Preliminary results showed that conditioning only on the original binary rewards is equivalent to foregoing return-to-go conditioning altogether. For further details, refer to \cref{app:appendix-f}.

\subsection{Baselines and variants} 
Following the Uni[MASK] approach~\citep{Carroll2022UniMASKUI}, we use the same base architecture for all our models. All model configurations are illustrated in Appendix~\ref{app:appendix-b}. We obtain IL baselines (where feedback is \textsc{None}) by masking all additional tokens except the one-off mission instructions. We unmask relevant additional tokens to achieve variants, specifically \emph{returns-to-go} for variant \textsc{Scalar} or \emph{language feedback} for \textsc{Lang}, as well as \textit{both} for \textsc{Combo}, and optionally predict the scalar reward or language feedback. For comprehensive details on training see \cref{app:appendix-g}. 


\subimport{tables}{compositionality}
\section{Results}
\subsection{Compositional Generalisation}
\paragraph{Systematicity.} We test if models can generalise to unseen combinations of goal colour and shape at test time. \cref{tab:compositionality} shows that performance using language feedback is comparable to the performance with scalar rewards when the model is trained to predict its own feedback, while combining language feedback and scalar rewards slightly surpasses the performance achieved with either feedback modalities when trained with the auxiliary feedback prediction objective. The performance of models trained with suboptimal trajectories and some form of feedback is more than four times higher than for the IL baselines.

\paragraph{Productivity.} In the context of our embodied setting, we define productivity as the ability to extrapolate to unseen values of mission or environment parameters. We isolate dimensions of compositionality in the mission in the form of categorical variables such as goal attributes, while in the environment, we identify compositionality w.r.t numerical variables, specifically the room dimensions or the number of obstacles. Note that while for compositionality in the mission, we use the language feedback and scalar rewards corresponding to the \textit{task feedback} oracle, we measure compositionality in the environment in the context of \emph{affordance feedback}. 

The results in \cref{tab:compositionality} indicate that language feedback is most effective when extrapolating to unseen goal colours, more complex layouts or higher numbers of obstacles. In these cases, language feedback performs roughly on par with the scalar rewards. In contrast, language feedback is less effective when tested on object locations that never appeared in the training data. We hypothesise that the models struggle to correctly ground language describing spatial relations and therefore, language feedback hinders more than it helps. We also test whether models trained on simple sequential instructions can extrapolate to more complex, compound instructions with multiple connectors. Specifically, models must understand the different connectors between two sequences (`and', `then', or `after'), which only occur in isolation in the training data. Surprisingly, learning from multiple trajectories per mission task improves performance of the baseline on unseen sequence tasks. It is conceivable that seeing successful sequences connected by `and' in different orders helps the model learn differences between order-agnostic coordinating conjunctions from temporal adverbs and subordinating conjunctions (`then' and `after'). However, we find that language feedback only provides comparable benefits to scalar rewards when are trained with the auxiliary feedback prediction task. 

\subimport{figures}{robustness}
\subsection{Robustness}
\label{sec:robustness}

\paragraph{Representations of subgoals.} We see evidence that models trained with language feedback or scalar rewards learn more robust representations of how to solve \textit{individual} subgoals. We find that models trained without any form of feedback partially complete many tasks but rarely complete \textit{all} subgoals required for task success, whereas those trained with feedback signals and several more or less optimal solutions for a given task appear to be significantly more successful at fully completing tasks, as illustrated in \cref{fig:robustness}a. 
Equipping models with language feedback and requiring them to learn how to predict their own feedback can further reduce the number of tasks that are only partially completed. Unlike the scalar rewards, language feedback reflects whether actions lead to \textit{subgoal} success or to \textit{task} success. Therefore, language feedback-conditioned models which are also trained to predict their own feedback can develop a more nuanced understanding of the difference between partial and full task completion. 

\paragraph{Robustness to lack of optimal data.} To simulate scenarios where optimal trajectories are categorically not available, we further test to what extent our method allows us to rely exclusively on suboptimal trajectories. \cref{tab:performance-drop} illustrates how,
when we omit the optimal trajectories, the advantage of the models that have access to some form of feedback becomes even clearer. We note the most pronounced relative performance drop for the baseline trained without step-level feedback (\textsc{None}), while removing the optimal data has the least impact on the \textsc{Combo} model trained on scalar and language feedback. This suggests that the step-level feedback effectively enables trajectory stitching from different suboptimal solutions.
\subimport{tables}{no-optimal}

\paragraph{Robustness to external 
perturbations.} We simulate external perturbations, such as hardware failure, using a variant of \textit{sticky actions} \citep{Machado2017RevisitingTA}. However, rather than delaying execution by $k$ steps, we simply randomly replace the current action with the sticky action. As a result, we can test the agent's ability to recover without inflating the number of steps taken on the environment. \cref{fig:robustness}b shows that models trained with suboptimal trajectories but without any form of feedback (\textsc{None w/ ST}) to contextualise mistakes are more the most adversely affected by external perturbations, with performance dropping to under 70\%. For models trained on language feedback, it is paramount to be able to predict feedback (\textsc{LANG + FP}), as this allows them to robustly anticipate the outcome of a given action, even if it is not the action that was originally predicted. This means the model can retain 75\% of the original performance, making it equally as robust to external perturbations as models trained with scalar rewards (\textsc{Scalar} and \textsc{SCALAR + FP}). Contextualising the language feedback with scalar rewards (\textsc{COMBO + FP}) can further aid recovery, leading to the most robust model variant, which retains 78\% of the original performance when faced with external perturbations. 

\paragraph{Robustness to adversarial or missing language feedback.} 
We test the sensitivity of models trained with language feedback to scenarios where feedback is either unhelpful or \textit{adversarial}, or entirely unavailable. We simulate the adversarial case by providing random English sentences at random time steps. \cref{fig:robustness}c-d show that in these settings, the performance of models  trained with language feedback collapses. While we consider this a positive sign that models do not simply learn to rely on spurious correlations \citep{Parekh2024InvestigatingTR}, it is ultimately desirable for them to be effective and safe in a range of deployment scenarios. Our experiments further reveal that when models are trained on both language and the corresponding scalar rewards and additionally required to learn to predict the resulting language feedback for their actions (\textsc{COMBO + FP}), they retain 70\%--80\% of their original performance. We attribute this effect specifically to the \textit{combination} of scalar rewards as context and the ability to anticipate feedback, as in isolation, these augmentations do not provide any tangible benefits. For additional adversarial test scenarios, refer to \cref{app:appendix-h}. 

\subimport{tables}{path-efficiency}
\paragraph{Solution efficiency.} 
We normalise the path lengths \textit{of successful episodes} by the corresponding oracle path lengths and use this as a proxy for how efficient solutions were. \cref{tab:path-efficiency} shows that all models, including the baseline that only learns from optimal trajectories, on average favour solutions that are less efficient than the oracle paths. At the same time, all models are able to find more optimal paths than the oracle\footnote{While the BFS-planner oracle always yields successful paths that are near-optimal but not guaranteed to be optimal.}, including for missions seen during training. As expected, including suboptimal trajectories but no feedback makes solutions the least efficient, while only learning from optimal trajectories results in the highest path efficiency. The results indicate that when models are trained with language feedback and the auxiliary feedback prediction task, they are able to learn the most efficient solutions from suboptimal data. 
\subimport{tables}{data-efficiency}
\subimport{figures}{data-efficiency}
\subsection{Data efficiency and data scaling}
We compare the performance on a representative compositional generalisation setting (\textit{Systematicity}) for all models, as well as a PPO \cite{Schulman2017ProximalPO} baseline\footnote{We adopt implementation and hyper-parameters in the original BabyAI benchmark, as detailed in \cref{app:appendix-c}.} when training on 25\%, 50\% and 100\% of the data used in our experiments. The results in \cref{fig:data-efficiency} suggest that the models contextualising suboptimal training trajectories with any form of feedback are not only more data efficient than either IL from optimal trajectories or PPO, they also show much stronger scaling properties. It appears that the \textsc{Combo} variant shows the strongest late-stage scaling, which suggests that this setting may require more data to outperform settings using single feedback types.


\section{Conclusion}
The continued popularity of language modelling objectives for LLMs sparked great interest in revisiting IL techniques for embodied agents (e.g., \citealp{ehsani2024spoc}). However, these are limited due to their inability to exploit suboptimal trajectories for learning more generalisable and robust behaviour. In this paper, we demonstrate the potential of language feedback as an efficient and intuitive feedback mechanism for IL when the dataset either contains or consists entirely of suboptimal trajectories, and show its potential as a viable alternative to scalar rewards for tasks specified in language. We define an experimental setup based on a highly configurable 2D grid-world environment for learning from pixel-based, partial observations, which we designed expressly to assess performance along different axes of compositional generalisation~\citep{Hupkes2019CompositionalityDH}. We show that models trained with both optimal and suboptimal data and feedback exhibit superior compositional generalisation abilities and increased robustness. We find that language feedback is not only a suitable alternative for scalar rewards for harnessing the learning potential of suboptimal samples, but also shows how we can increase robustness of models trained with language feedback with auxiliary feedback or by combining the strengths of both approaches. We are optimistic that our findings will inform future work on LLM-based policies and that language feedback can play a role in different fine-tuning regimes and for alignment with human preferences.

\section*{Limitations}

\paragraph{Environment complexity.} This work is conducted in BabyAI-XGen, a grid-world environment that, while providing controlled experimental conditions, falls short of the visual realism, continuous action spaces, and physics fidelity of real-world robotic tasks. We chose this environment to enable precise control over a range of compositional generalization factors, and to ensure reliable trajectory generation, which would be difficult to achieve in more complex 3D environments. However, the central principle of learning from suboptimal trajectories contextualized with feedback is domain-agnostic, and future work should explore scaling to more realistic environments where optimal trajectories may be even harder to obtain.

\paragraph{Limited model architecture.} Our approach uses a relatively small Transformer backbone (\textasciitilde{90M} parameters) trained from scratch, which may not reflect the capabilities of larger foundation models. This choice was made to avoid confounding factors from pre-trained knowledge and ensure fair comparison across feedback conditions. Future work should investigate how pre-trained language models and the corresponding larger architectures might better leverage feedback signals, particularly for more complex reasoning about multi-step errors and corrections \cite{ChiyahGarcia2024RepairsIA}, keeping in mind that careful scaling of training data to match the increased model capacity will likely be required, as our current dataset size could lead to overfitting or suboptimal utilization of larger architectures.

\paragraph{Synthetic feedback generation.} We generate feedback programmatically using environment internals rather than studying feedback quality or generation mechanisms. This approach enables controlled experimentation without confounding factors from inconsistent or incorrect model-generated feedback while minimising computational overheads. However, future work should explore the development of more sophisticated step-level feedback generation mechanisms involving VLM-based systems.

\paragraph{Cost and practicality of feedback collection.} While our results demonstrate the effectiveness of language feedback, practical deployment faces important cost considerations. We focused on systematic feedback generation to isolate the core research question of whether feedback can transform suboptimal data into useful learning signals. However, for many embodied domains, step-level scalar rewards and language feedback will likely require human annotation or sophisticated reward models, with language feedback potentially incurring higher costs than scalar rewards. Future work should systematically evaluate these tradeoffs across different domains and explore efficient feedback collection strategies, particularly for interactive domains where language feedback is inherently more intuitive than scalar scores—such as collaborative robotics requiring natural human-robot communication, interactive tutoring systems where explanatory feedback aids learning, and multi-agent coordination tasks where agents need to understand and communicate about errors and corrections.


\section*{Potential Risks and Ethical Considerations}

We believe our work on enhancing embodied AI through constructive language feedback holds significant promise for developing more robust and adaptable robotic systems. By enabling agents to learn from a wider range of demonstrations, including suboptimal ones, we move closer to creating AI that can effectively operate in complex and unpredictable real-world environments.
However, the ability of AI agents to learn from and act upon language feedback also introduces important ethical considerations and potential risks. As these systems become more sophisticated and integrated into our lives, ensuring their safety, reliability, and fairness becomes paramount. For instance, if language feedback is biased or malicious, an embodied AI agent could learn and perpetuate harmful behaviors or make decisions with unintended negative consequences in real-world scenarios. Aware of this potential issue, we have designed our experimental setup with robustness in mind and defined specific scenarios where we are challenging models with unexpected actions or adversarial language instructions. As shown in our experiments, models trained to predict feedback are more resilient to these perturbations, offering a more compelling solution for future applications.

Due to these reasons, we also decided to focus on a controllable environment where action execution can be simulated without major repercussions on the world and the humans within it. However, we also acknowledge that, due to this reason, the models trained in this setting might be biased towards simulation environments and might not directly generalize to real-world environments.


\clearpage
\bibliography{bib}


\renewcommand{\thetable}{\Alph{section}.\arabic{table}}
\renewcommand{\thefigure}{\Alph{section}.\arabic{figure}}

\clearpage
\appendix

\setcounter{table}{0}
\setcounter{figure}{0}
\subimport{appendix}{appendix-a}
\setcounter{table}{0}
\setcounter{figure}{0}
\subimport{appendix}{appendix-b}
\setcounter{table}{0}
\setcounter{figure}{0}
\subimport{appendix}{appendix-c}
\setcounter{table}{0}
\setcounter{figure}{0}

\subimport{appendix}{appendix-d}
\setcounter{table}{0}
\setcounter{figure}{0}
\subimport{appendix}{appendix-e}
\setcounter{table}{0}
\setcounter{figure}{0}
\subimport{appendix}{appendix-f}
\setcounter{table}{0}
\setcounter{figure}{0}
\subimport{appendix}{appendix-g}
\setcounter{table}{0}
\setcounter{figure}{0}
\subimport{appendix}{appendix-h}

\setcounter{table}{0}
\setcounter{figure}{0}
\subimport{appendix}{appendix-i}

\end{document}

%% file: content/math_commands.tex
\def\eqref#1{equation~\ref{#1}}

\def\1{\bm{1}}

\DeclareMathAlphabet{\mathsfit}{\encodingdefault}{\sfdefault}{m}{sl}
\SetMathAlphabet{\mathsfit}{bold}{\encodingdefault}{\sfdefault}{bx}{n}



%% file: figures/overview.tex
\begin{figure*}[tb]
    \centering
    \includegraphics[width=\linewidth]{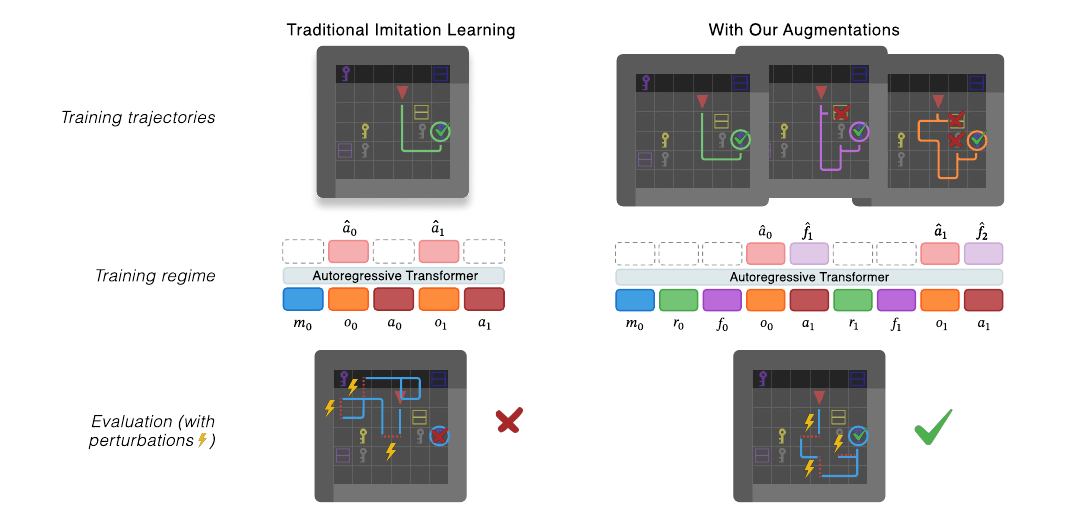}
    \caption{Our method leverages both optimal and suboptimal trajectories for a given task instance by contextualising modes of behaviour with feedback signals. We leverage different types of feedback and additional self-supervised auxiliary tasks to learn highly generalisable and robust representations of behaviour in a data-efficient manner. }
    \label{fig:overview}
\end{figure*}

%% file: figures/masking-strategies-shortened.tex
\begin{figure}[tb]
    \centering
    \includegraphics[width=0.9\linewidth]{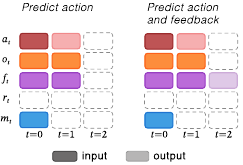}
    \caption{Input and output tokens for a model conditioning action generation on initial instructions and language feedback, with the option to predict language feedback at the next time step. $m_i$=instructions, $f_i$=language feedback, $r_i$=returns-to-go/rewards, $o_i$=observations, $a_i$=actions. }
    \label{fig:masking-strategies-shortened}
\end{figure}

%% file: figures/mixed-trajectories.tex
\begin{figure}[tb]
    \centering
    \includegraphics[width=\linewidth]{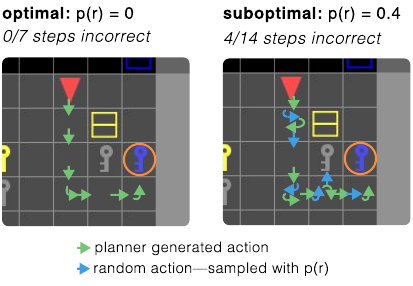}
    \caption{Optimal trajectories generated by a planner, and suboptimal trajectories obtained by replacing planner actions with random actions, which the planner must correct if necessary. The given \code{p(r)} is exemplary.}
    \label{fig:mixed-trajectories}
\end{figure}

%% file: tables/compositionality.tex
\begin{table*}[tb]
\centering
\footnotesize
\begin{tabular}{l cc c ccccc c}
\toprule
\multirow{3}{*}{Feedback} & \multicolumn{2}{c}{Enhancement} & \multicolumn{1}{c}{Systematicity} & \multicolumn{5}{c}{Productivity} \\
\cmidrule(lr){2-3} \cmidrule(lr){4-4} \cmidrule(lr){5-9}
& ST & FP & Color-Shape & Color$^\dagger$ & Location$^\dagger$ & Sequence$^\dagger$ & Layout$^*$ & Obstacles$^*$ & All \\
\midrule
\multirow{2}{*}{\textsc{None}} 
& $\times$ & $\times$ & 16.9 & 13.5 & 4.9 & 10.0 & 70.3 & 82.2 & 33.1 \\
& $\checkmark$ & $\times$ & 15.1 & 14.1 & 2.8 & 44.4 & 75.1 & 83.1 & 39.1 \\
\midrule
\multirow{2}{*}{\textsc{Scalar}}
& $\checkmark$ & $\times$ & 69.6 & 68.0 & 19.7 & 69.8 & 80.6 & 87.3 & 65.8 \\
& $\checkmark$ & $\checkmark$ & 69.4 & 68.8 & 21.3 & 71.9 & 80.0 & 85.6 & 66.2 \\
\midrule
\multirow{2}{*}{\textsc{Lang}}
& $\checkmark$ & $\times$ & 68.0 & 66.5 & 13.8 & 59.1 & 80.9 & 88.3 & 62.8 \\
& $\checkmark$ & $\checkmark$ & 69.7 & 69.1 & 14.6 & 66.1 & 79.2 & 85.9 & 64.1 \\
\midrule
\multirow{2}{*}{\textsc{Combo}}
& $\checkmark$ & $\times$ & 70.3 & 66.6 & 17.7 & 67.3 & 78.0 & 83.9 & 64.0 \\
& $\checkmark$ & $\checkmark$ & 71.7 & 68.6 & 20.3 & 69.5 & 78.5 & 81.6 & 65.0 \\
\bottomrule
\end{tabular}
\caption{Success rates (\%) across different dimensions of compositional generalisation: Systematicity (combinatorial interpolation to novel combinations) and Productivity (extrapolating to unseen values). We average the performance across multiple tasks, each requiring both navigation and object interaction/manipulation. A breakdown of tasks per generalization setting can be found in Table E.1 in the appendix. $^\dagger$compositionality in the mission, $^*$compositionality in the environment. ST=suboptimal trajectories, FP=feedback prediction.}
\label{tab:compositionality}
\end{table*}

%% file: figures/robustness.tex
\begin{figure*}[tb]
    \centering
    \includegraphics[width=\linewidth]{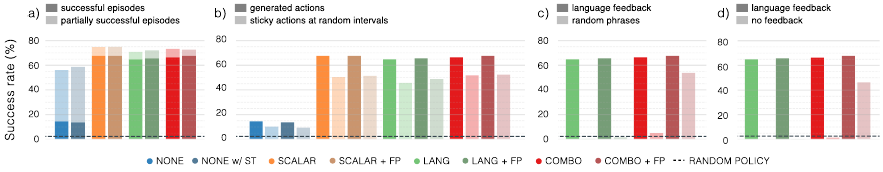}
    \caption{Comparison of success rates under various robustness evaluation settings. We report robustness results on in-distribution data and average over Pickup and PutNext tasks. From left to right: a) representations of subgoals, b) external perturbations, c) adversarial feedback, d) missing feedback. b)-d) correspond to alternative inference scenarios. ST=suboptimal trajectories, FP=feedback prediction.}
    \label{fig:robustness}
\end{figure*}

%% file: tables/no-optimal.tex
\begin{table}[tb]
\centering
\small
\setlength{\tabcolsep}{4pt}
\begin{tabular}{l cc c}
\toprule
Feedback & ST + OT & ST only & Drop \\
\midrule
\textsc{None} & 15.1 & 12.1 & -19.9\% \\
\textsc{Scalar} & 69.6 & 62.3 & -10.5\% \\
\textsc{Lang} & 68.0 & 61.2 & -10.0\% \\
\textsc{Combo} & 70.3 & 65.6 & -6.7\% \\
\bottomrule
\end{tabular}
\caption{Relative drop in success rate (\%) when using only suboptimal trajectories (ST) vs suboptimal + optimal trajectories (ST+OT). Results are averaged for in-distribution Pickup and PutNext missions.}
\label{tab:performance-drop}
\end{table}

%% file: tables/path-efficiency.tex
\begin{table}[tb]
\centering
\small
\setlength{\tabcolsep}{5pt}
\begin{tabular}[width=\linewidth]{@{\extracolsep{\fill}}l cc SSS@{}}
\toprule
& \multicolumn{2}{c}{Enhancement} & \multicolumn{3}{c}{ON Path Lengths} \\
\cmidrule(lr){2-3} \cmidrule(lr){4-6}
Feedback & ST & FP & \multicolumn{1}{c}{Mean} & \multicolumn{1}{c}{Min} & \multicolumn{1}{c}{Min (train)} \\
\midrule
\multirow{2}{*}{\textsc{None}} 
& $\times$ & $\times$ & {220.9\%} & {94.1\%} & {91.7\%} \\
& $\checkmark$ & $\times$ & {253.8\%} & {75.0\%} & {89.5\%} \\
\midrule
\multirow{2}{*}{\textsc{Scalar}}
& $\checkmark$ & $\times$ & {244.1\%} & {69.2\%} & {86.2\%} \\
& $\checkmark$ & $\checkmark$ & {239.1\%} & {69.2\%} & {89.5\%} \\
\midrule
\multirow{2}{*}{\textsc{Lang}}
& $\checkmark$ & $\times$ & {240.8\%} & {69.2\%} & {89.5\%} \\
& $\checkmark$ & $\checkmark$ & {233.6\%} & {75.0\%} & {89.5\%} \\
\midrule
\multirow{2}{*}{\textsc{Combo}}
& $\checkmark$ & $\times$ & {245.4\%} & {69.2\%} & {86.2\%} \\
& $\checkmark$ & $\checkmark$ & {236.1\%} & {75.0\%} & {89.5\%} \\
\bottomrule
\end{tabular}
\caption{Oracle-normalised path lengths of successful episodes for in-distribution missions, averaged over Pickup and PutNext tasks. <100\% corresponds to more efficient solutions, and >100\% to less efficient solutions. ST=suboptimal trajectories, FP=feedback prediction.}
\label{tab:path-efficiency}
\end{table}

%% file: tables/data-efficiency.tex
\begin{table}[tb]
\centering
\small
\setlength{\tabcolsep}{4pt}
\begin{tabular}[width=\linewidth]{@{\extracolsep{\fill}}l cc ccc@{}}
\toprule
& \multicolumn{2}{c}{Enhancement} & \multicolumn{3}{c}{Proportion of data} \\
\cmidrule(lr){2-3} \cmidrule(lr){4-6}
Feedback & ST & FP & 25\% & 25\%→50\% & 50\%→100\% \\
\midrule
\multirow{2}{*}{\textsc{None}} 
& $\times$ & $\times$ & 13.1 & +0.6 & +3.3 \\
& $\checkmark$ & $\times$ & 11.6 & +3.0 & +0.5 \\
\cmidrule{1-6}
\multirow{2}{*}{\textsc{Scalar}}
& $\checkmark$ & $\times$ & 53.6 & +10.0 & +6.0 \\
& $\checkmark$ & $\checkmark$ & 55.0 & +9.6 & +4.9 \\
\cmidrule{1-6}
\multirow{2}{*}{\textsc{Lang}}
& $\checkmark$ & $\times$ & 50.7 & +9.5 & +7.8 \\
& $\checkmark$ & $\checkmark$ & 52.0 & +7.7 & +10.0 \\
\cmidrule{1-6}
\multirow{2}{*}{\textsc{Combo}}
& $\checkmark$ & $\times$ & 53.4 & +4.6 & +12.2 \\
& $\checkmark$ & $\checkmark$ & 49.9 & +11.1 & +10.8 \\
\bottomrule
\end{tabular}
\caption{Change in success rate (\%) when doubling the amount of training data. ST=suboptimal trajectories, FP=feedback prediction.}
\label{tab:data-efficiency-deltas}
\end{table}

%% file: figures/data-efficiency.tex
\begin{figure}[tb]
    \centering
    \includegraphics[width=\linewidth]{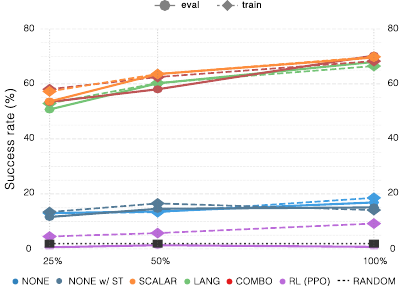}
    \caption{Change in success rate on \textit{Systematicity} tasks as the amount of training data increases, averaged over Pickup and PutNext tasks. ST=suboptimal trajectories}
    \label{fig:data-efficiency}
\end{figure}

%% file: appendix/appendix-a.tex
\section{Further background on related work}
\label{appendix-a}

\paragraph{Language-guided embodied AI } An exception to the customary approach of creating embodied AI datasets from planner-generated trajectories is TEACh \citep{Padmakumar2021TEAChTE}, which was collected from pairs of human annotators collaborating to complete tasks in the simulated environment. A proportion of TEACh trajectories contain inefficiencies and corrections, typically resulting from miscommunication between the  \citep{Min2022DontCT}. However, any such occurrences of suboptimal behaviour in TEACh are incidental rather than systematic, and the dataset does not include multiple example trajectories for the same task instance.

\paragraph{Language as feedback}
For the purpose of our study, we consider language feedback from an oracle that is external to the agent itself, and therefore does not directly apply to cases where agents self-improve through an internal process of self-reflection~\cite{Yao2022ReActSR, madaan2024self}.

\paragraph{Compositional generalisation}\label{app:appendix-b}
As has been argued in \citet{Kirk2021ASO}, many of the categories of compositional generalisation originally introduced by \citet{Hupkes2019CompositionalityDH} for language are applicable for agents in interactive environments; we focus on two of the five categories: \emph{Systematicity} and \emph{Productivity} and test to what extent learning from multiple solutions and suboptimal behaviour can translate into increased Systematicity and Productivity in agents that have access to different feedback signals. For details on the remaining three categories, we refer the reader to the original definitions in \citet{Hupkes2019CompositionalityDH}.
In the context of embodied agents and interactive environments, Systematicity pertains to the ability to systematically recombine known components and rules to novel combinations \citep{Hupkes2019CompositionalityDH, Kirk2021ASO}. \citet{Kirk2021ASO} refer to this as \emph{combinatorial interpolation}, as the agent needs to interpolate to values of environment parameters which it has seen independently but not in combination. Productivity, which \citet{Hupkes2019CompositionalityDH} define as the ability of models to generate to output sequences that exceed the length of the sequences seen during training, is loosely equivalent to \emph{extrapolation} in \citep{Kirk2021ASO}, according to which the values for a single or multiple environment parameters fall outside the ranges seen during training; as the resulting environments tend to be more complex, agents are typically required to generate longer trajectories than those present in the input. In the case of language-guided embodied tasks, we can define compositionality with respect to the language of the mission instructions or the complexity of the environment.

%% file: appendix/appendix-b.tex
\section{Method details}\label{app:appendix-b}

\subimport{../tables}{llama-config}

\subimport{../figures/appendix}{masking-strategies}

\paragraph{Transformer backbone.} Unlike the Decision Transformer, which uses a scaled-down GPT-2 model\cite{Radford2019LanguageMA} at its core, we adapt architecture from Llama2 \citep{Touvron2023LLaMAOA} to serve as our Transformer backbone, which allows us to leverage innovations such as the RoPE (Rotary Position Embedding) positional encodings, Grouped-Query Attention (GQA), and longer context lengths to facilitate learning from entire trajectories, rather than sub-trajectories of limited context lengths, as well as more efficient training. Specifically, we use the Flash Attention implementation of Llama from Huggingface. Note that we do not use the original Llama2 configuration or weights; instead we use the model dimensions corresponding to those use for the Base size of Octo \citep{Team2024OctoAO}, a pretrained Transformer-based robot policy, and randomly initialise weights. We give an overview of the configuration of the Llama-style backbone in \cref{tab:llama-config}

\paragraph{Flexible token masking.} Similar to \citet{Carroll2022UniMASKUI}, we devise a flexible masking scheme to facilitate conditioning on different additional tokens while keeping the model architecture constant and reusing the same datasets for different feedback options. Supported additional tokens range from one-off language instructions, to scalar rewards, language feedback, and their combination. Tokens are either masked out for the whole sequence if irrelevant for a given test case or only for those time steps they were not provided for. 
\cref{fig:masking-strategies-a} illustrates the masking schemes used for our models used in this study, while \cref{fig:masking-strategies-b} highlights additional use cases of our masking schemes. For instance, we can accommodate arbitrary combinations of feedback prediction objectives, and optionally repeat the mission instructions at every step

\paragraph{Token embeddings.} While our design is deliberately modular and supports a range of image and text encoder options, we opt for sentence-level text embeddings from a fully frozen Sentence-BERT \citep{Reimers2019SentenceBERTSE} corresponding to the pre-trained general purpose model \code{all-mpnet-base-v2}, for the mission instructions and language feedback (where applicable) to allow us to condense long text inputs into compact vector representations corresponding to a single token, and a custom CNN image encoder (see \cref{tab:cnn_architecture}) proposed for pixel-based RL experiments in MiniGrid and BabyAI by \citep{willems2023rl}, which we train concurrently with the policy, and which condenses the image input into one token. Compressing the multi-modal input modalities into a single token per timestep most closely aligns with the input into the Transformer backbone is constructed in Decision Transformers. We project language embeddings using a projection layer similar to LlaVa \citep{Liu2023VisualIT}, 
where $f(\mathbf{X}_t)$ is the text encoder, and $\mathbf{W}$ is the projection matrix which maps the text features into the target embedding space with the dimensionality of the Llama-style backbone (\cref{eq:text-projection}). The image embeddings are projected using a simple projection layer in keeping with the implementation of the CNN used for BabyAI.
\subimport{../equations}{text-projection}
\subimport{../tables}{cnn-architecture}

We embed the discrete actions and timesteps using Embedding layers. After adding the timestep embeddings to the token-specific embeddings, the token embeddings for each timestep are interleaved to form flattened sequences, $ (m_1,r_1,f_1,o_1,a_1, ...\hspace{0.15em}, m_T,r_T,f_T,o_T,a_T) $, where $T$ is equal to the number of steps of the trajectory. The interleaved token embeddings are then normalised and passed as input to the Llama-style backbone, along with the corresponding attention mask, where we mask out unused tokens. Note that, unlike in the original Decision Transformer, we leverage an additional level of position embeddings for each token in the flattened sequence in the Llama-style backbone.

\paragraph{Prediction heads and losses.} 
We follow previous work on Decision Transformers~\cite{Chen2021DecisionTR} and learn to predict actions in the environment as a next token prediction task by minimising cross-entropy loss of the actions. We consider this as our main loss function for the training of the agent's policy. We consider several auxiliary losses aimed at predicting property tokens at the next timesteps. We define auxiliary losses for a range of property tokens including \textit{image tokens}, \textit{reward tokens}, and \textit{feedback tokens}. Our losses are inspired by the latent imagination method by \citet{Elliott2017ImaginationIM}. For each loss, we leverage the hidden state $\mathbf{h}_t$ associated with the timestep $t$ of our Transformer backbone to predict the embedding of the property token at the next timestep $\mathbf{p}_{t+1}$ via a dedicated prediction head $P$. We report below the details of each loss.

\paragraph{Action token prediction.} For the action prediction, we treat this as a classification task over the discrete action space. We assume the target to be the action $a_{t+1}$. We define a prediction head $P_a$ as a linear layer that takes the hidden state $\mathbf{h}_t$ and outputs unnormalized logits over the action vocabulary. We use the cross-entropy loss as follows:
\subimport{../equations}{action-loss-ce}

\paragraph{Feedback token prediction.} For the feedback prediction, we assume that our embedding representation for the feedback is the Sentence-BERT representation $\mathbf{f}_{t+1}$ associated with the feedback in position $t+1$. We define a prediction head $P_f$ as a linear layer with \textsc{GeLU} activations. We use the MSE loss as follows:
\subimport{../equations}{feedback-loss-mse}

\paragraph{Reward token prediction.} For the reward prediction, we assume the target to be the reward $G_{t+1}$ associated with the reward in position $t+1$ (see \cref{app:appendix-e} for details). We define a prediction head $P_r$ as a linear layer with \textsc{Sigmoid} activation. We use the MSE loss as follows:
\subimport{../equations}{reward-loss-mse}

\paragraph{Image token prediction.} For the image prediction, we follow a similar definition to \citet{Elliott2017ImaginationIM}, but instead of the MSE loss, we calculate the loss based on the Cosine similarity between the image embedding the $\mathbf{i}_{t+1}$ produced by the CNN encoder for the timestep $t+1$, and the image embeddings predicted by the prediction head consisting of a linear layer $P_i$ with \textsc{ReLU} activation. We use a Cosine loss as follows:
\subimport{../equations}{image-loss-cosine}

In our experiments, we found this loss did not improve performance consistently compared to the baseline (see \cref{app:appendix-g}). We leave for future work exploring more specific training regimes which can better leverage this loss following work in self-supervised representation learning~\cite{jaderberg2016reinforcement, caron2021emerging}.

\paragraph{Weighted average loss.} Due to the difference in magnitude between the different losses, we learn via SGD a loss-specific weight to balance the contribution of each auxiliary loss. We compute the final loss as a weighted average of the different losses used by a certain model configuration.

\subimport{../tables}{model-architecture}

%% file: tables/llama-config.tex
\begin{table}
\centering
\begin{tabular}{ll}
\toprule
Parameter & Value \\
\midrule
vocab\_size & \textit{1} \\
hidden\_size & \textit{768} \\
intermediate\_size & \textit{3072} \\
num\_hidden\_layers & \textit{768} \\
num\_attention\_heads & \textit{12} \\
num\_key\_value\_heads & \textit{12} \\
hidden\_act & SiLU \\
max\_position\_embeddings & 2048 \\
initializer\_range & 0.02 \\
rms\_norm\_eps & 1e-6 \\
use\_cache & True \\
pad\_token\_id & None \\
bos\_token\_id & 1 \\
eos\_token\_id & 2 \\
pretraining\_tp & 1 \\
tie\_word\_embeddings & False \\
rope\_theta & 10000.0 \\
rope\_scaling & None \\
attention\_bias & False \\
attention\_dropout & 0.0 \\
mlp\_bias & False \\
head\_dim & None \\
\bottomrule
\end{tabular}
\caption{Configuration of our Llama-style backbone. Most values are taken from the huggingface implementation \citep{huggingface_llama2}. Values we override are highlighted in italics.}
\label{tab:llama-config}
\end{table}

%% file: figures/appendix/masking-strategies.tex
\begin{figure}[tbhp]
    \centering
    \begin{subfigure}{\columnwidth}
        \centering
        \includegraphics[width=0.9\columnwidth]{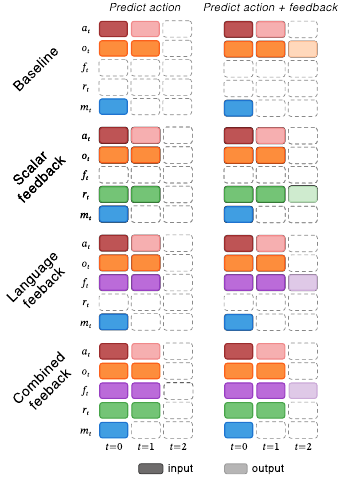}
        \caption{Input and output tokens of the models used in our experiments. For instance, models using \emph{scalar feedback} condition action generation on instructions and returns-to-go, and optionally predict rewards of the next time step.}
        \label{fig:masking-strategies-a}
    \end{subfigure}
    
    \vspace{1em}
    
    \begin{subfigure}{\columnwidth}
        \centering
        \includegraphics[width=\columnwidth]{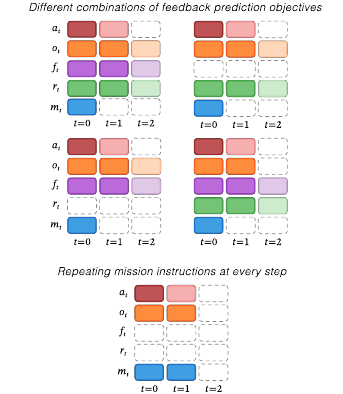}
        \caption{Further possible models supported by our flexible token masking scheme.}
        \label{fig:masking-strategies-b}
    \end{subfigure}
    
    \caption{We use a flexible token masking scheme inspired by \citet{Carroll2022UniMASKUI} to achieve different configurations from the same base model to mask unused tokens, where $m_i$=instructions, $f_i$=language feedback, $r_i$=returns-to-go/rewards, $o_i$=observations, $a_i$=actions.}
    \label{fig:masking-strategies}
\end{figure}

%% file: equations/text-projection.tex
\begin{equation}
    \label{eq:text-projection}
       \mathbf{H}_t = \mathbf{W} \cdot \mathbf{Z}_t, \text{ with } \mathbf{Z}_t = f(\mathbf{X}_t)
\end{equation}

%% file: tables/cnn-architecture.tex
\begin{table}[tb]
\centering
\small
\setlength{\tabcolsep}{4pt}  
\begin{tabular}{lcccc}
\toprule
Block & Operation & In Ch. & Out Ch. & Kernel \\
\midrule
Conv Block 1 & Conv2D & 3 & 16 & (2, 2) \\
& ReLU & 16 & 16 & -- \\
\midrule
Pooling & MaxPool2D & 16 & 16 & (2, 2) \\
\midrule
Conv Block 2 & Conv2D & 16 & 32 & (2, 2) \\
& ReLU & 32 & 32 & -- \\
\midrule
Conv Block 3 & Conv2D & 32 & 64 & (2, 2) \\
& ReLU & 64 & 64 & -- \\
\midrule
Output & Flatten & 64 & -- & -- \\
\bottomrule
\end{tabular}
\caption{Architecture of the custom CNN used to embed image observations adopted from \citep{willems2023rl}.}
\label{tab:cnn_architecture}
\end{table}

%% file: equations/action-loss-ce.tex
\begin{align}
    \mathcal{L}_a = -\sum_{t=1}^{T} \log(p(a_t | \mathbf{h}_{t-1}))
\end{align}

%% file: equations/feedback-loss-mse.tex
\begin{equation}
    \label{eq:feedback-loss-mse}
    \mathcal{L}_f = \text{MSE}(P_f(\mathbf{h}_t), \mathbf{f}_{t+1})
\end{equation}

%% file: equations/reward-loss-mse.tex
\begin{equation}
    \label{eq:reward-loss-mse}
    \mathcal{L}_r = \text{MSE}(P_r(\mathbf{h}_t), G_{t+1})
\end{equation}

%% file: equations/image-loss-cosine.tex
\begin{equation}
    \label{eq:image-loss-cosine}
    \mathcal{L}_i = 1 - \cos(P_i(\mathbf{h}_t), \mathbf{i}_{t+1})
\end{equation}

%% file: tables/model-architecture.tex
\begin{table}[h]
\centering
\small
\setlength{\tabcolsep}{4pt}
\begin{tabular}{lll}
\toprule
Name & Type & Params \\
\midrule
\emph{Embed returns} & \emph{Linear} & \emph{1.5K} \\
Embed images & Custom CNN & 10.5K \\
Embed actions & Embedding & 5.4K \\
Embed timesteps & Embedding & 444K \\
\midrule
Project image embeddings & Linear & 30.7M \\
& ReLU & \\
\midrule
Project language embedding & Linear & 1.2M \\
& Linear & \\
& GELU & \\
\midrule
Normalise embeddings & LayerNorm & 1.5K \\
\midrule
Backbone & LlamaModel & 113M \\
\midrule
Predict actions & Linear & 4.6K \\
\emph{Predict feedback emb.} & \emph{Linear} & \emph{590K} \\
& \emph{GELU} & \\
\emph{Predict image emb.} & \emph{Linear} & \emph{590K} \\
& \emph{ReLU} & \\
\emph{Predict rewards} & \emph{Linear} & \emph{769} \\
& \emph{Sigmoid} & \\
\bottomrule
\end{tabular}
\caption{Model architecture, for a total parameter count of $\sim$146M. Optional modules in italics. Note that we pass only the instructions through the language embedding projection layers for models that don't use language feedback. Details on the CNN and the Llama backbone are provided separately.}
\label{tab:model_architecture}
\end{table}

%% file: appendix/appendix-c.tex
\section{PPO baseline
}\label{app:appendix-c}
The PPO baseline is trained on the equivalent amount of data as our multi-task IL models but only on single tasks (e.g. Pickup \textit{or} PutNext). This corresponds to \textasciitilde9K, \textasciitilde18K and \textasciitilde36K trajectories, and evaluated only on the equivalent evaluation task. We train five PPO models corresponding to the same global seeds used for our other models, and average their performance. Note that \citet{ChevalierBoisvert2018BabyAIAP} find that solving the comparable \verb|PickupLoc| and \verb|PutNext| tasks from the original BabyAI suite \texttt{using symbolic observations} requires \textasciitilde1.4-1.6M and \textasciitilde2.2-2.7M training episodes, respectively. Our findings show that the generalisation performance of the PPO baseline when trained on only up to 36K trajectories, corresponding to less than \textasciitilde2.5\% of data, is marginally below random performance. We use the hyper-parameters and acrhitecture corresponding to the implementation from the GitHub repository published for \citet{ChevalierBoisvert2018BabyAIAP}\footnote{\href{https://github.com/mila-iqia/babyai}{github.com/mila-iqia/babyai}}, but use the pixel-based observations and the corresponding image encoder.

%% file: appendix/appendix-d.tex
\section{Environment details}\label{app:appendix-d}

\subimport{../figures/appendix}{babyaix-parameters}

BabyAI \citep{ChevalierBoisvert2018BabyAIAP} builds on top of MiniGrid to procedurally generate so-called \emph{missions} in abstract 2D grid worlds, where a mission is defined as the combination of a set of mission instructions and initial environment state \citep{ChevalierBoisvert2018BabyAIAP}.\footnote{We use a similar definition to describe the more generic term of task instance.} The discrete action space consists of navigation and manipulation actions (\verb|left|, \verb|right|, \verb|forward|, \verb|pickup|, \verb|drop|, \verb|toggle|) and an optional \verb|done| action. As we do not require the agent to indicate when it has finished a task, we constrain the action space to exclude the \verb|done| action, which reduces the number of possible actions to $\lvert X \rvert = 6$. By default, BabyAI environments are partially observable, and observations are symbolic representations of a top-down view of the grid. BabyAI approximates an egocentric agent perspective by masking out tiles in the grid that are currently not in the agent's field of view because they are hidden behind walls or behind the agent. Since we desire models to learn from pixels, we use an RGB image wrapper to map partial observations into pixel space.

The 19 original levels in BabyAI were designed to be used as a multi-task benchmarking suite and for curriculum learning, and evaluates agents on a range of skills. Besides understanding action language and goal object attributes, agents need to be able to navigate single rooms and multi-room mazes, unblock the path to doors, and guess to unlock doors during navigation. Additionally, some levels test the agent's understanding of spatial or temporal language: the agent may have to identify goal objects as specified by a location descriptor (e.g., `pick up the blue key on your left'), or to understand composite instructions, whereby multiple instruction clauses are chained together in a sequence that may or may not require the instructions to be executed in a specific order (e.g., `pick up the green ball after you pick up the purple box').

As BabyAI was not designed with compositional generalisation in mind, it does not expose the means to control environment parameters such that $p_{train} \neq p_{test}$. Additionally, due to the mechanism by which instructions and objects are sampled to procedurally generate missions, not all missions for a given level will test the skill introduced by the level. From initial experimentation, we find that understanding location language and composite instructions, as well as unblocking or unlocking doors, and even maze navigation are not strictly required for all missions of those levels meant to test the respective skill. While we assume that this is a deliberate design choice allowing a phasing in of skills for curriculum learning, it limits the level of control that can be exerted over the parameters of the environment further, and hinders the systematic evaluation of different skills.

As \citet{Kirk2021ASO} argue, environments should be controllable and possess a structured parameter space to constitute a suitable test bed for most forms of compositional generalisation. As is the case with BabyAI, controllability is typically not supported for environments that facilitate procedural generation. Previous work attempting to use BabyAI for compositional generalisation \citep{McCallum2023IsFA} achieves different training and test distributions by generating a large number of BabyAI missions and filtering for those with the desired goal object attributes. This approach scales poorly to larger datasets and beyond one or two parameters. We devise a more scalable approach and build on top of the BabyAI-MiniGrid ecosystem to develop \textsc{BabyAI=XGen}, a version of BabyAI which gives users full control over a comprehensive range of environment parameters by simply passing a \code{config} object with the desired configuration when instantiating an environment with \code{gym.make}. We provide an exhaustive list of controllable parameters in \cref{tab:controllable-parameters}, and illustrate a selection of parameters in \cref{fig:babyai-x}. As BabyAI is now integrated into Gymnasium~\cite{towers2024gymnasium}, we will release \textsc{BabyAI-XGen} following Apache 2.0 License.

\subimport{../tables}{controllable-parameters}

%% file: figures/appendix/babyaix-parameters.tex
\begin{figure}[tb]
    \centering
    \includegraphics[width=\linewidth]{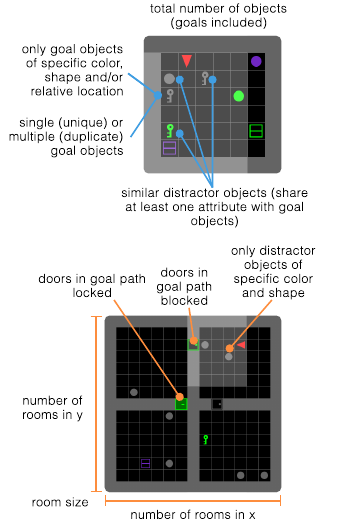}
    \caption{An illustration of selected controllable parameters in our version of BabyAI.}
    \label{fig:babyai-x}
\end{figure}

%% file: tables/controllable-parameters.tex
\begin{table*}[htbp]
\centering
\small
\begin{tabular}{p{0.3\textwidth}p{0.65\textwidth}}
\toprule
\textbf{Parameter} & \textbf{Description} \\
\midrule
\texttt{room\_size} & The number of cells in a room \\
\texttt{num\_rows} & How many rooms a maze should have (in the y dimension) \\
\texttt{num\_cols} & How many rooms the maze should have (in the x dimension) \\
\texttt{remove\_walls} & Whether to combine multiple maze rooms by removing walls \\
\midrule
\texttt{num\_objs} & The total number of objects in a given room or maze (including goals and distractors/obstacles) \\
\texttt{duplicate\_goals} & Whether to include multiple duplicate instances of the goal object, or only one unique goal object; affects article usage in instructions (e.g., ``pickup the blue key'' vs ``pickup a blue key'') \\
\texttt{dists\_unique} & Whether all distractor objects should be unique \\
\texttt{dists\_include\_similar} & Whether to include distractor objects that share attributes with goal objects; adds a distractor for each specified goal object attribute (color, shape, location) \\
\texttt{avoid\_overlapping\_goals} & Whether to avoid sampling goal objects with overlapping attributes \\
\midrule
\texttt{only\_objs} & Permissible object attributes as lists of \texttt{(color, shape)} combinations for distractors and goals \\
\texttt{only\_locations} & Similar to \texttt{only\_objects} but for lists of \texttt{location} \\
\texttt{exclude\_objects} & Disallowed object attributes as lists of \texttt{(color, shape)} combinations \\
\texttt{exclude\_locations} & Similar to \texttt{exclude\_objects} but for lists of \texttt{location} \\
\midrule
\texttt{distinguish\_by} & Whether to refer to goal objects by color, shape and/or location in instructions; \texttt{False} means never use that attribute, \texttt{True} enables random usage unless specified in \texttt{strict} \\
\texttt{action\_kinds} & The action language available in mission instructions: can be one or multiple of \texttt{goto}, \texttt{open}, \texttt{pickup}, or \texttt{putnext} \\
\texttt{instr\_kinds} & Whether to generate single instructions (\texttt{action}) or sequential instructions (\texttt{and}, \texttt{before}, \texttt{after}), or combinations \\
\texttt{seq\_complexity} & Number of single instructions in sequences: \texttt{low} (two), \texttt{medium} (three), \texttt{high}, or combinations \\
\texttt{multiple\_locations} & Controls location language for multiple goal objects: \texttt{True}, \texttt{'strictly'}, or \texttt{False} \\
\midrule
\texttt{unblocking} & Controls path unblocking requirements: \texttt{False}, \texttt{True}, or \texttt{'strictly'} (maze-only) \\
\texttt{explicit\_unlocking} & Whether goal doors need unlocking as additional goal condition (maze-only) \\
\texttt{implicit\_unlocking} & Similar to \texttt{unblocking}, but for unlocking doors along goal path \\
\texttt{all\_doors\_open} & Whether all maze doors start open (maze-only) \\
\texttt{goal\_room\_same\_as\_start} & Controls goal object placement relative to start room: \texttt{False}, \texttt{True}, or \texttt{'strictly'} (maze-only) \\
\midrule
\texttt{verify} & Optional verification settings: \texttt{'use\_done\_action'} or \texttt{'strict'} mode \\
\bottomrule
\end{tabular}
\caption{Configuration parameters for maze generation and mission instructions}
\label{tab:controllable-parameters}
\end{table*}

%% file: appendix/appendix-e.tex
\section{Dataset details
}\label{app:appendix-e}
For the suboptimal trajectories, we use $p_1=0.5$ and $p_2=0.75$ to replace planner actions with random actions and achieve different degrees of suboptimality (see \cref{eq:suboptimal-actions}. While we refer to the planner generated trajectories as \emph{optimal}, we acknowledge that BFS does not guarantee optimality; however, BFS is efficient and there is no exploration. 
\subimport{../equations}{suboptimal-actions}

We construct a range of multi-task datasets with an equal number of trajectories per task, where the number of training tasks ranges from three for the experiments for compositionality in language, to nine for the experiments testing compositionality in the environment. We list the of the training and evaluation tasks, along with details on the configurations used to instantiate the tasks, in \cref{tab:tasks}. As \citep{Kirk2021ASO} note, a model that has seen a wide range of possible values for a parameter during training will be able to perform better on unseen values at test time than a model whose training data contained only one possible value for the parameter. We therefore ensure that we include examples of multiple possible parameter values in the training data. Likewise, we expose models to missions requiring skills of various degrees of difficulty, whereby harder skills build on top of harder skills (\verb|Goto| < \verb|Pickup| < \verb|PutNext|). For mazes, we assume a similar relationship between single rooms and mazes (where single room < maze), and open, closed and locked doors (where open doors < closed doors < locked doors).

\subimport{../tables}{tasks}

%% file: equations/suboptimal-actions.tex
\begin{equation}
    \label{eq:suboptimal-actions}
    A_t = \begin{cases}
        \begin{array}{ll}
            a, & \text{with prob. } 1-\rho, \\
            x \sim \mathcal{U}(\mathcal{A}), & \text{with prob. } \rho
        \end{array}
    \end{cases}
\end{equation}

%% file: tables/tasks.tex
\begin{table*}
\small
\centering
\setlength{\tabcolsep}{5pt}
\begin{tabular}{lllp{5.5cm}}
\toprule
\multicolumn{4}{l}{\textbf{Compositionality in Instructions}} \\
\midrule
& Train Datasets & Test Datasets & Important Configurations \\
\midrule
\multicolumn{4}{l}{\textbf{Systematicity}} \\
& GoToColorType & GoToColorTypeUnseen & Train: Excludes yellow balls from goals and distractors \\
& PickupColorType & PickupColorTypeUnseen & Test: Yellow balls as goals only, with similar distractors maintained \\
& PutNextColorType & PutNextColorTypeUnseen1 & \\
& & PutNextColorTypeUnseen2 & \\
\midrule
\multicolumn{4}{l}{\textbf{Productivity (color)}} \\
& GoToOnlyColor & GoToOnlyColorUnseen & Train: Excludes yellow objects from goals and distractors \\
& PickupOnlyColor & PickupOnlyColorUnseen & Test: Yellow objects as goals only, with similar distractors maintained \\
& PutNextOnlyColor & PutNextOnlyColorUnseen1 & \\
& & PutNextOnlyColorUnseen2 & \\
\midrule
\multicolumn{4}{l}{\textbf{Productivity (location)}} \\
& GoToLoc & GoToLocUnseen & Train: Excludes 'left' location from goals \\
& PickupLoc & PickupLocUnseen & Test: Only includes 'left' location for goals \\
& PutNextLoc & PutNextLocUnseen & \\
& & PutNextLocUnseen2 & \\
\midrule
\multicolumn{4}{l}{\textbf{Productivity + Systematicity (seq)}} \\
& PickupEasySeqAnd & PickupMediumSeqBefore & Train: Only obstacles as distractors, with low sequence complexity \\
& PickupEasySeqBefore & PickupMediumSeqAfter & Test: Only obstacles as distractors, with medium sequence complexity \\
& PickupEasySeqAfter & & \\
\midrule
\multicolumn{4}{l}{\textbf{Compositionality in Environment}} \\
\midrule
\multicolumn{4}{l}{\textbf{Productivity}} \\
& Pickup & PickupN16 & Base: Only includes obstacles as distractors \\
& PickupMaze1X2 & PickupMaze2X3 & Regular: Goal in different room, all doors open \\
& PickupMaze1X2Blocked & PickupMaze2X3Blocked & Blocked: Goal in different room, requires moving obstacles \\
& PickupMaze1X2DoorsClosed & PickupMaze2X3DoorsClosed & Closed: Goal in different room, doors initially closed \\
& PickupMaze1X2DoorsLocked & PickupMaze2X3DoorsLocked & Locked: Goal in different room, requires finding and using keys \\
& PickupMaze2X2 & PickupMaze3X3 & \\
& PickupMaze2X2Blocked & PickupMaze3X3Blocked & \\
& PickupMaze2X2DoorsClosed & PickupMaze3X3DoorsClosed & \\
& PickupMaze2X2DoorsLocked & PickupMaze3X3DoorsLocked & \\
\bottomrule
\end{tabular}
\caption{Overview of environment configurations used to generate our multitask training datasets and test environments. Note that for our robustness experiments (see \cref{sec:robustness}), we use the training configurations but instantiate the test environments with unseen seeds.}
\label{tab:tasks}
\end{table*}

%% file: appendix/appendix-f.tex
\section{Feedback details
}\label{app:appendix-f}

\subimport{../figures/appendix}{task-feedback-template}

To provide granular language feedback and denser scalar rewards for all conceivable tasks in our custom BabyAI, we extend the rule-based feedback oracles proposed in \citet{McCallum2023IsFA}.

\paragraph{Task feedback.} In line with \citep{Xi2024TeachingER}, we increase the informativeness of the language feedback compared with \citet{McCallum2023IsFA} and construct the feedback string according to an updated template, increasing diversity by sampling from a range of predefined options when populating the generic parts of the feedback template shown in \cref{fig:task-feedback-template}. 

\paragraph{Affordance feedback.} We adopt the \emph{rule feedback} oracle from \citet{McCallum2023IsFA} without any changes, but will refer to such language feedback, which is triggered when agents execute actions that have no visible effect on the environment due to its affordances, as \emph{affordance feedback}. Such `failed' actions include, for instance, the agent trying to move forward through a closed door, trying to open a wall, or attempting to put an object on top of another object (which is not permitted in BabyAI). A number of examples of task feedback and affordance feedback are provided in the appendix.

\paragraph{Shaped rewards.}  We leverage the success and action failure detection mechanisms in the feedback oracles to return intermediate scalar rewards. These shaped rewards replace the default reward function in BabyAI, which only provides a sparse reward signal in the form of terminal, binary rewards as shown in 
\cref{eq:babyai-default-reward-function}, where $\gamma$ is the discount factor, $n$ the number of steps the agent took, and $N_T$ the step budget for a given task in BabyAI.
\subimport{../equations}{babyai-default-reward-function}. 
This was informed by initial experiments showed only marginal differences between the IL case and return-to-go conditioning with the original binary rewards (see \cref{tab:sparse-rewards}), as the returns-to-go are identical for every timestep except the last. Furthermore, they are identical for all training trajectories, since we are not including trajectories for failed episodes. 
\subimport{../tables}{sparse-rewards}
According to the revised reward function, agents receive a fraction of the terminal reward based on the total number of subgoals in the task each time they complete a subgoal; for each failed action, we assign a small negative reward. The negative action failure rewards can be combined with a binary terminal reward, or the dense subgoal rewards.

Note that the target return during testing is set to the terminal reward ($G=1.0$).

\subimport{../equations}{babyai-updated-reward-function}
\subimport{../equations}{babyai-updated-reward-function-2}
\subimport{../equations}{babyai-updated-reward-function-3}

%% file: figures/appendix/task-feedback-template.tex
\begin{figure*}[tbph]
    \centering
    \includegraphics[width=\linewidth]{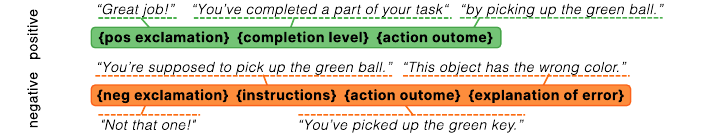}
    \caption{We extend the task feedback oracle in \citet{McCallum2023IsFA} and provide positive and negative task feedback with templates. We sample exclamations from predefined options, the rest is specific to mission and action.}
    \label{fig:task-feedback-template}
\end{figure*}

%% file: equations/babyai-default-reward-function.tex
\begin{equation}
    \label{eq:babyai-default-reward-function}
    R(T) = \begin{cases} 1 - \gamma \cdot \frac{n}{N_T} & \text{\small if task success} \\ 0 & \text{\small otherwise} \end{cases}
\end{equation}

%% file: tables/sparse-rewards.tex
\begin{table}[tb]
\centering
\begin{tabular}{lc}
\toprule
Model & Success Rate (\%) \\
\midrule
\textsc{None} & 12.7 \\
\textsc{Scalar (binary)} & 13.7 \\
\textsc{Scalar (shaped)} & 69.5 \\
\bottomrule
\end{tabular}
\caption{Success rate (\%) when using no feedback (\textsc{None}), sparse returns-to-go (\textsc{Scalar (binary)}) and dense returns to go (\textsc{Scalar (shaped)})}
\label{tab:sparse-rewards}
\end{table}

%% file: equations/babyai-updated-reward-function.tex
\begin{equation}
    \label{eq:babyai-updated-reward-function}
    R(G) = \begin{cases}
    \frac{1}{|G|} & \text{\small if subgoal achieved} \\
    0 & \text{\small otherwise}
    \end{cases}
\end{equation}

%% file: equations/babyai-updated-reward-function-2.tex
\begin{equation}
    \label{eq:babyai-updated-reward-function-2}
    R(F) = \begin{cases}
    -0.01 & \text{\small if failed action} \\
    0 & \text{\small otherwise}
    \end{cases}
\end{equation}

%% file: equations/babyai-updated-reward-function-3.tex
\begin{equation}
    \label{eq:babyai-updated-reward-function-3}
    R(T) = \begin{cases}
    1 & \text{\small if task success} \\
    0 & \text{\small otherwise}
    \end{cases}
\end{equation}

%% file: appendix/appendix-g.tex
\section{Experimental framework}\label{app:appendix-g}

\paragraph{Baselines and variants.} Our flexible architecture as described in \cref{sec:method} serves as the basis for all our baselines and variants; we simply unmask additional token modalities for \emph{scalar feedback} (returns-to-go) and/or \emph{language feedback} and optionally predict additional tokens, specifically the scalar rewards and language feedback. For all model types, we initialise five models using different random seeds. We use the same hyperparameters across models in the same model family, that is, models that use the same feedback signal; note that preliminary hyperparamter-focused experiments, show that slightly different batch sizes and learning rates are best for the baseline, the variant using language feedback, and the variants using scalar rewards. The hyperparameter choices can be found in \cref{tab:hyperparameters}. Note that rather than sub-trajectories of a given context length, all models use the full trajectory as context; trajectories are padded to the length of the longest trajectory in the batch. Despite trying different model sizes following \citet{Team2024OctoAO}, we report results of the \textsc{base} size for all our model variants.

\subimport{../tables}{hyperparameters}

\paragraph{Computing infrastructure and computational budget.} We trained and evaluated all models using up to six NVIDIA A40 GPUs with 48GB, with runs taking around 3-8 hours when training on multi-task datasets with three tasks, and approximately 25 hours when training on multi-task datasets spanning nine tasks. Evaluation runs took between 1.5 and 7.5 hours, depending on the complexity and number of evaluation tasks, as well as the quality of the obtained models\footnote{Models with poor performance took longer to evaluate than models with fair or good performance, as performance translates into the number of evaluation episodes that lasted for the maximum number allowed steps}. In a total, the required compute budget for this work given the hardware used is ~2,000 GPU hours.

%% file: tables/hyperparameters.tex
\begin{table}[tb]
\centering
\begin{tabular}{ll}
\hline
Hyperparameter & Value \\
\hline
Optimizer & AdamW \\
Weight Decay & 0.0001 \\
Gradient Clip Threshold & 0.25 \\
Learning Rate & 1e-5* \\
Batch Size & 32*  \\
Max training epochs & 20 \\
Early stopping min delta & 0.01 \\
Early stopping monitor & action loss (val) \\
Early stopping mode & min \\
Early stopping patience & 2 epochs \\
Total trajectories seen & 33,684 ** \\
Examples per optimizer step & 128 \\
\hline
\end{tabular}
\caption{Hyperparameters using during model training. *For the variant with the language feedback (\textsc{Language}), we use lr=2.5e-5, and for the variants using scalar feedback (\textsc{Scalar} and \textsc{Combined}), we use lr=5e-5 and a batch size of 64. **For the experiments testing compositionatily in the environment, models will have seen 110,052 trajectories. Note that the length of the trajectories varies quite significantly, and we refrain from providing an estimate for the number of transitions seen.}
\label{tab:hyperparameters}
\end{table}

%% file: appendix/appendix-h.tex
\section{Further experimental details and results}\label{app:appendix-h}
\subimport{../figures/appendix}{replace-feedback}
\subimport{../figures/appendix}{strict-success}

All results are averaged across 128 missions per evaluation task and 5 different model seeds. Note that we use designated evaluation configurations for our compositional generalisation experiments; for all other experiments, we instantiate the environments according to the training configurations but with unseen seeds to ensure that agents have to generalise to unseen task instances, while disentangling the results from skills required for compositional generalisation.

\paragraph{External perturbations.}
We adapt \emph{sticky actions} from \citet{Machado2017RevisitingTA}, who introduced this setting to introduce non-deterministic behaviour in the Atari Learning Environment \citep{Bellemare2012TheAL}. At every time step, instead of the agent's current action, the environment executes the agent's previous action again with a probability defined by stickiness parameter $\zeta$. Concretely, at time step $t$ the agent chooses an action $a$ and takes a step in the environment with action ${A}_{t}$. The full equation is given in the appendix.
\subimport{../equations}{sticky_actions}  \ref{eq:sticky-actions} 
According to this, sticky actions may be executed for $k$ consecutive time steps with a probability of $(1-\zeta)^k\zeta$. 

\paragraph{Adversarial feedback.} We test in total three adversarial scenarios, whereby we: 1) provide random English sentences at random timesteps, or replace language feedback with 2) random Lorem Ipsum sentences, or 3) random English sentences. Note that for English sentences, we avoid sampling words which occur in the actual language feedback. We consider case 1), the results for which are reported in \cref{sec:robustness}, the most adversarial scenario, as feedback is neither semantically relevant nor timely. A similar setting to 2) and 3) has been explored in \citet{Parekh2024InvestigatingTR}, who perturb instructions into \textit{Gobbledygook Words}. 
Instead, we explicitly choose adversarial feedback that is similar to the actual feedback. We show cases 2) and 3) in \cref{fig:replace-feedback}, and can identify no noticeable difference for the baseline and the models trained on both language feedback and scalar feedback between the two cases, and only minimal difference for the models trained only on language feedback. We note that the performance of the most robust model is slightly worse when feedback is missing (see \cref{fig:robustness}d), compared with the adversarial cases.

\paragraph{Solution Efficiency.} 
We calculate our oracle-path normalised path lengths as shown in Equation \cref{eq:on-paths}, where $L_i^\ast$ is the number of steps in the expert demonstration for a given task instance, and $\hat{L_i}$ the number of steps it took the model to complete the task, averaged across all successful episodes. This is inspired by the path-weighted success rate in \citet{Shridhar2019ALFREDAB}, which combines the overall success rate with the efficiency of successful paths. We choose to disentangle these two aspects of success and provide them as separate metrics, specifically success rate and partial success rate across all evaluation episodes, and oracle-path normalised path lengths only for successful episodes.
\subimport{../equations}{on-paths}

\paragraph{Strict Task Success.}
As an additional setting, we test whether models trained with language feedback learn behaviours that see them unnecessarily interacting with distractor objects to elicit feedback, which may be undesirable during deployment and results in less efficient solutions. Using the \emph{strict} evaluation setting, which terminates episodes early as failed when the agent interacts with (picks up) distractor objects. \cref{fig:strict-success} shows that when evaluating models in \emph{strict} mode, performance does not deteriorate noticeable for models trained with access to feedback signals, while the performance of the baseline drops to 70\% of the original performance when trained only on optimal trajectories, and 65\% when trained on both optimal and suboptimal trajectories, which indicates that the baseline exhibits more inefficient trial-and-error behaviour at inference.

\subimport{../tables}{image-loss}
\paragraph{Image prediction loss.} We find that models that can predict the embedding of the next image observation, which could be considered \emph{visual feedback}, do not tend to outperform models that only predict the next action (see \cref{tab:image-loss}). We hypothesize that this is in part due to the abstract nature of the grid world and the birds eye view, where there is little variation in the observations, with the background and objects being represented as solid colour blocks. Since objects take up at most one cell and in combination with the discrete actions, object interactions will never change more than 2 cells (or 3\% of the visible pixels assuming an 8x8 room) when the agent is moved forward by one cell, or 1 cell when an object disappears or reappears as a result of a pickup or drop action, and can be as minimal as a few pixels when the triangle of the agent is rotated 90 degrees as a result of a left or right navigation action. Conversely, due to the partially observable nature of the environment and the discrete nature of the action space, navigation actions, particularly turning left or right, may result in a considerable part of the environment abruptly changing from visible to obscured or vice versa, whereby the cells that are not visible to the agent will simply be masked out. A more realistic observation space with a first person view and more fine-grained actions may be a better fit for testing the potential of leveraging the next observation as visual feedback similar to human learning.

\paragraph{Model scale.} To validate our architectural choices, we conduct experiments examining how model capacity affects performance across different feedback modalities. We train models of varying sizes—tiny (\textasciitilde{10}M), small (\textasciitilde{30}M), and base (\textasciitilde{90}M) parameters—while keeping training data constant. The results, shown in \cref{tab:model-scaling}, demonstrate that performance plateaus beyond a certain capacity threshold across all feedback conditions. These findings indicate that BabyAI-XGen presents fundamental challenges that cannot be addressed through increased model capacity alone. The performance bottleneck appears to stem from the compositional generalization requirements rather than insufficient model capacity. Beyond the tiny-to-small transition, additional parameters provide diminishing returns, suggesting that the core challenge lies in learning compositional reasoning patterns rather than requiring larger representational capacity. This behavior aligns with the lottery ticket hypothesis \citep{Frankle2018TheLT}, which demonstrates that overparameterized networks contain sparse subnetworks achieving comparable performance, indicating that much additional parameter space remains unused for the specific task requirements. In our compositional generalization setting, the fundamental difficulty lies in developing reasoning capabilities for novel task combinations rather than memorizing complex patterns that would benefit from increased capacity. Our approach prioritizes isolating the fundamental mechanisms of learning from suboptimal trajectories with feedback. While pretrained language models represent an important direction for future work—particularly regarding how pretraining might qualitatively affect feedback processing—our current architecture provides sufficient capacity to demonstrate the effectiveness of our approach while maintaining experimental clarity and avoiding confounding factors from pretrained knowledge.

\subimport{../tables/}{model-scaling}

%% file: figures/appendix/replace-feedback.tex
\begin{figure}[tb]
    \centering
    \includegraphics[width=\linewidth]{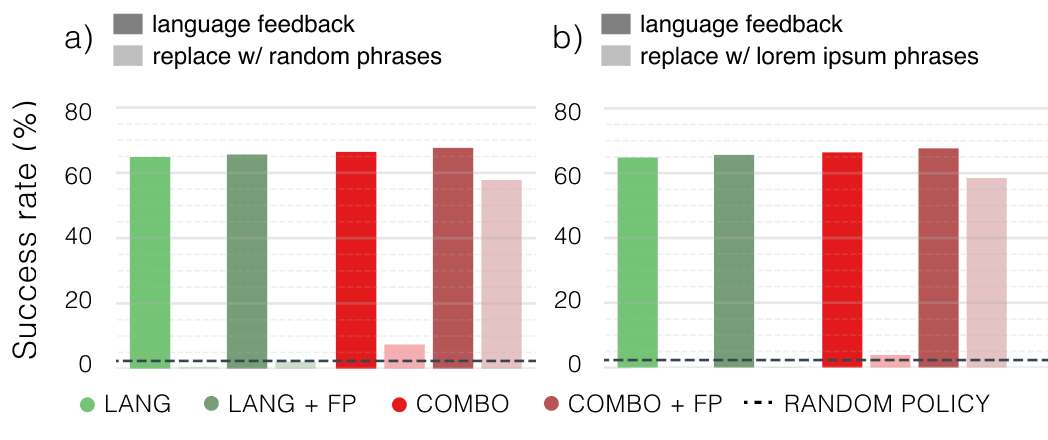}
    \caption{Comparison of success rates under additional adversarial language feedback evaluation settings. We average performance on in-distribution Pickup and PutNext tasks. From left to right: a) replace feedback with random sentences, d) replace feedback with lorem ipsum sentences. ST=suboptimal trajectories, FP=feedback prediction}
    \label{fig:replace-feedback}
\end{figure}

%% file: figures/appendix/strict-success.tex
\begin{figure}[tb]
    \centering
    \includegraphics[width=\linewidth]{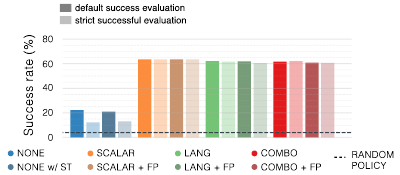}
    \caption{Comparison of success rates when interactions with non-goal objects are allowed vs. when they lead to early termination (task failure). We average performance on in-distribution Pickup and PutNext tasks. Note this refers to the inference scnenario. ST=suboptimal trajectories, FP=feedback prediction.}
    \label{fig:strict-success}
\end{figure}

%% file: equations/sticky_actions.tex
\begin{equation}
    \label{eq:sticky-actions}
    A_t = \begin{cases}
        a, \text{ with prob. } 1-\zeta, \\
        a_{t-1}, \text{ with prob. } \zeta.
    \end{cases}
\end{equation}

%% file: equations/on-paths.tex
\begin{equation}
    \label{eq:on-paths}
    p_on = \frac{1}{N} \sum_{i \in \mathcal{S}}^{N} \frac{L_i^\ast}{\hat{L_i}}
\end{equation}

%% file: tables/image-loss.tex
\begin{table}[tb]
\centering
\begin{tabular}{lc}
\toprule
Model & Success Rate (\%) \\
\midrule
\textsc{None} & 14.1 \\
\textsc{None (FP)} & 14.4 \\
\bottomrule
\end{tabular}
\caption{Success rate (\%) of the baseline trained without any additional feedback (\textsc{None}) signal with and the auxiliary task of predicting the next image (FP).}
\label{tab:image-loss}
\end{table}

%% file: tables/model-scaling.tex
\begin{table}[tb]
\centering
\small
\begin{tabular}{l ccc}
\toprule
Feedback & Tiny (\textasciitilde{10}M) & Small (\textasciitilde{30}M) & Base (\textasciitilde{90}M) \\
\midrule
\textsc{None} & 11.9\% & 14.5\% & 12.7\% \\
\textsc{Scalar} & 64.7\% & 68.4\% & 69.6\% \\
\textsc{Lang} & 60.9\% & 67.6\% & 67.7\% \\
\textsc{Combo} & 67.6\% & 69.1\% & 70.4\% \\
\bottomrule
\end{tabular}
\caption{Success rate (\%) across different model sizes, showing plateauing beyond a certain capacity threshold.}
\label{tab:model-scaling}
\end{table}

%% file: appendix/appendix-i.tex
\section{Reproducibility}\label{app:appendix-i}
To meet the high reproducibility standards in the ML research community, a fully reproducible training and evaluation framework is available via \href{https://github.com/sabraaap/fossil}{github.com/sabraaap/fossil}. This includes implementations for all experiments carried out. Models were trained using PyTorch (Ansel et al., 2024) and Lightning (Falcon and The PyTorch Lightning Team, 2024), and dependencies were tracked using Poetry. Experiments were managed via Hydra configuration files (Yadan, 2019), and all configurations, commands, hyperparameters, and seeds used are available and signposted clearly in the provided codebase.

\paragraph{Environment.} We register BabyAI-XGen, our controllable environment based on BabyAI developed for generalisation research, as a Gymnasium environment and provide several predefined configuration objects to reproduce not only the environments used to generate our datasets and evaluate our models, but also blueprints for versions of the existing BabyAI level suite. This can be accessed at \href{https://github.com/sabraaap/fossil}{github.com/sabraaap/fossil}.

\paragraph{Training data.} All training data has been made available via Hugging Face Datasets repositories \href{https://huggingface.co/fossil-eai/datasets}{huggingface.co/fossil-eai/datasets}. In addition, training data can be regenerated using the dataset generation scripts which is available in the provided codebase at \href{https://github.com/sabraaap/fossil}{github.com/sabraaap/fossil}.

\paragraph{Model checkpoints.} All model checkpoints are provided on the Hugging Face Hub at \href{https://huggingface.co/fossil-eai/models}{huggingface.co/fossil-eai/models} to facilitate further experiments and explorations.